\documentclass[10pt,twocolumn,letterpaper]{article}

\usepackage{cvpr}              

\usepackage{graphicx}
\usepackage{amsmath}
\usepackage{amssymb}
\usepackage{booktabs}

\usepackage{paralist}

\usepackage{makecell}
\usepackage{subcaption}
\usepackage{multirow}
\usepackage{pifont}

\usepackage{bbm}
\usepackage{mathtools}
\usepackage[table]{xcolor}

\usepackage[pagebackref,breaklinks,colorlinks]{hyperref}

\usepackage[capitalize]{cleveref}
\crefname{section}{Sec.}{Secs.}
\Crefname{section}{Section}{Sections}
\Crefname{table}{Table}{Tables}
\crefname{table}{Tab.}{Tabs.}

\newcommand{\modelname}{\text{M2I}}
\newcommand{\modelnamespace}{\text{M2I} }

\newcommand{\rev}[1]{{#1}}

\def\cvprPaperID{10019} 
\def\confName{CVPR}
\def\confYear{2022}

\begin{document}

\title{M2I: From Factored Marginal Trajectory Prediction to Interactive Prediction}

\author{Qiao Sun$^{1}$\thanks{Denotes equal contribution. Code and demo available at paper website: \url{https://tsinghua-mars-lab.github.io/M2I/}  \texttt{alan.qiao.sun@gmail.com},\; \texttt{xhuang@csail.mit.edu}.} \quad Xin Huang$^{2\hspace{1pt}*}$\thanks{X. Huang was supported in part by Qualcomm Innovation Fellowship.} \quad Junru Gu$^{1}$ \quad Brian C. Williams$^{2}$ \quad Hang Zhao$^{1}$\thanks{Corresponding at: \texttt{hangzhao@mail.tsinghua.edu.cn}.} \\
        $^{1}$IIIS, Tsinghua University \quad $^{2}$CSAIL, Massachusetts Institute of Technology
}
\maketitle

\begin{abstract}
Predicting future motions of road participants is an important task for driving autonomously in urban scenes. Existing models excel at predicting marginal trajectories for single agents, yet it remains an open question to jointly predict scene compliant trajectories over multiple agents. The challenge is due to exponentially increasing prediction space as a function of the number of agents. In this work, we exploit the underlying relations between interacting agents and decouple the joint prediction problem into marginal prediction problems. Our proposed approach M2I first classifies interacting agents as pairs of influencers and reactors, and then leverages a marginal prediction model and a conditional prediction model to predict trajectories for the influencers and reactors, respectively. The predictions from interacting agents are combined and selected according to their joint likelihoods. Experiments show that our simple but effective approach achieves state-of-the-art performance on the Waymo Open Motion Dataset interactive prediction benchmark.
\end{abstract}

\vspace{-6mm}
\section{Introduction}
\vspace{-2mm}
\label{sec:intro}
Trajectory prediction is widely used by intelligent driving systems to infer future motions of nearby agents and identify risky scenarios to enable safe driving. Recent advances have shown great success in predicting accurate trajectories by learning from real-world driving examples. Many existing trajectory prediction works \cite{lee2017desire,chai2019multipath,tang2019multiple,gao2020vectornet,liang2020learning,gu2021densetnt} focus on generating marginal prediction samples of future trajectories over individual agents, failing to reason about their interactions in the future. As a result, the prediction samples over multiple agents may overlap with each other and result in sub-optimal performance.

\begin{figure}[t!]
    \centering
    \includegraphics[width=0.47\textwidth]{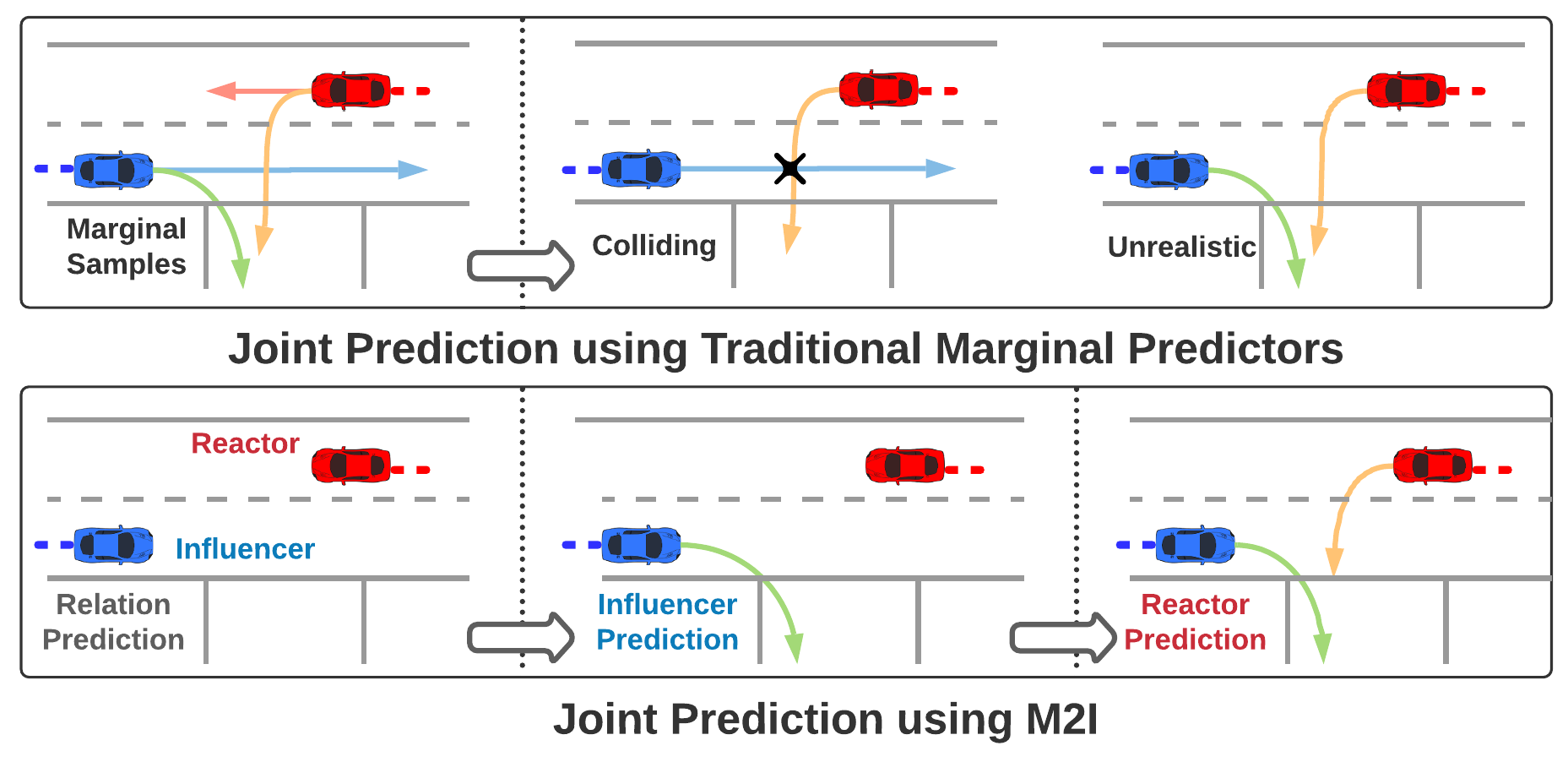}
    \vspace{-4mm}
    \caption{A motivating example of \modelname. Top: Traditional marginal predictor often produces \emph{scene inconsistent} trajectory predictions that collide with each other. Even for non-colliding predictions, it ignores the potential interaction between agent futures and may predict unrealistic behaviors. Bottom: Our proposed approach \modelname~predicts scene compliant trajectories by first identifying an influencer reactor pair in the scene. It then predicts marginal trajectories for the influencer and reactive trajectories for the reactor.}
    \label{fig:motivating_example}
\vspace{-6mm}
\end{figure}

We present a motivating example in \cref{fig:motivating_example}, in which a marginal predictor produces a set of prediction samples separately for two interacting agents, as visualized in the top left figure. While the predictions for each agent are reasonable without considering the presence of the other, some trajectory pairs will collide when considering them jointly. For instance, it is unlikely that the red agent turns left while the blue agent goes forward, as indicated in the top middle example in \cref{fig:motivating_example}. Therefore, it is necessary to predict scene compliant trajectories with the existence of multiple agents to support better prediction accuracy.

To generate scene compliant trajectories, one can learn a joint predictor to predict trajectories in a joint space over multiple agents; however, it suffers from an exponentially increasing prediction space as the number of agents increases. As investigated by~\cite{gu2021densetnt}, while it is feasible to predict a set of goals for a single agent, the goal space increases exponentially with the number of agents and becomes unmanageable for even two agents with a few hundred goal candidates for each agent. 
A more computationally efficient alternative to producing scene compliant trajectories is to post-process marginal prediction samples by pruning colliding ones; however, such an ad-hoc approach fails to take into account potential agent interactions in the future and may ignore other conflicts which are hard to prune by heuristics. For instance, although the prediction sample in the top right figure in \cref{fig:motivating_example} is collision-free, the red agent may slow down when turning left to keep a safe distance from the blue agent. Such an interactive behavior is hard to be captured by a marginal predictor as it is unaware of the future behavior of the other agents in the scene.

In this paper, we propose \modelnamespace that leverages marginal and conditional trajectory predictors to efficiently predict scene compliant multi-agent trajectories, by approximating the joint distribution as a product of a marginal distribution and a conditional distribution. The factorization assumes two types of agents: the \emph{influencer} that behaves independently without considering the other agents, and the \emph{reactor} that reacts to the behavior of the influencer. This assumption is inspired by the recent study on the underlying correlations between interactive agent trajectories~\cite{tolstaya2021identifying}.  Under the assumption, we leverage a standard marginal predictor to generate prediction samples for the influencer, and a conditional predictor to roll out future trajectories for the reactor conditioned on the future trajectory of the influencer. The advantage of our proposed approach \modelname~is illustrated in the bottom figures in \cref{fig:motivating_example}, in which we first predict the relations of the interactive agents. Given the relations, we predict the future trajectories of the influencer and then predict reactive behaviors of the reactor conditioned on each influencer prediction. 
As causality in driving interaction remains an open problem\cite{tolstaya2021identifying}, we pre-label the influencer-reactor relation based on a heuristic, and propose a relation predictor to classify interactive relations at inference time. 

Our contributions are three-fold. 
First, we propose a simple but effective framework \modelnamespace that leverages marginal and conditional predictors to generate accurate and scene compliant multi-agent trajectories. The framework does not assume a specific predictor structure, allowing it to be adopted by a wide range of backbone prediction models. 
Second, we propose a relation predictor that infers high-level relations among interactive agents to decouple the prediction space. 
Third, we demonstrate our framework using a goal-conditioned prediction model. Experiments show that \modelnamespace achieves state-of-the-art performance on the Waymo Open Motion Dataset interactive prediction benchmark.

\vspace{-2mm}
\section{Related Work}
\vspace{-1mm}
Trajectory prediction for traffic agents has been studied extensively in recent years. Due to uncertainty in human intent, the future trajectories are probabilistic and multi-modal. To handle the multi-modality problem, 
\cite{chai2019multipath,salzmann2020trajectron++} propose models that output behavior predictions as Gaussian mixture models (GMMs), in which each mixture component represents a single modality. Instead of parameterizing the prediction distribution, generative models, such as generative adversarial models (GANs) \cite{huang2020diversitygan,gupta2018social,zhao2019multi} and (conditional) variational autoencoders (VAEs) \cite{salzmann2020trajectron++,mangalam2020pecnet,yuan2019diverse,lee2017desire}, produce trajectory samples to approximate the distribution space. These generative models suffer from sample inefficiency and require many samples to cover diverse driving scenarios \cite{huang2020diversitygan}.

More recently, a family of models are proposed to improve prediction accuracy and coverage by first predicting high-level intentions, such as goal targets  \cite{rhinehart2019precog,zhao2020tnt,mangalam2020pecnet,fang2020tpnet,gilles2021gohome,gu2021densetnt}, lanes to follow~\cite{song2021learning,kim2021lapred}, and maneuver actions \cite{deo2018multi,dai2020flexible,huang2021hyper,kuo2022trajectory}, before predicting low-level trajectories conditioning on the intention. Such models demonstrate great success in predicting accurate trajectories for single agents in popular trajectory prediction benchmarks, such as Argoverse~\cite{chang2019argoverse} and Waymo Open Motion Dataset~\cite{ettinger2021large}. While our proposed approach \modelnamespace can use an arbitrary prediction model, we choose to adopt an anchor-free goal-based predictor~\cite{gu2021densetnt} because of its outstanding performance.

In the rest of the section, we introduce the literature closely related to our approach, on interactive trajectory prediction and conditional trajectory prediction.

\begin{figure*}[t!]
    \centering
    \includegraphics[width=0.8\textwidth]{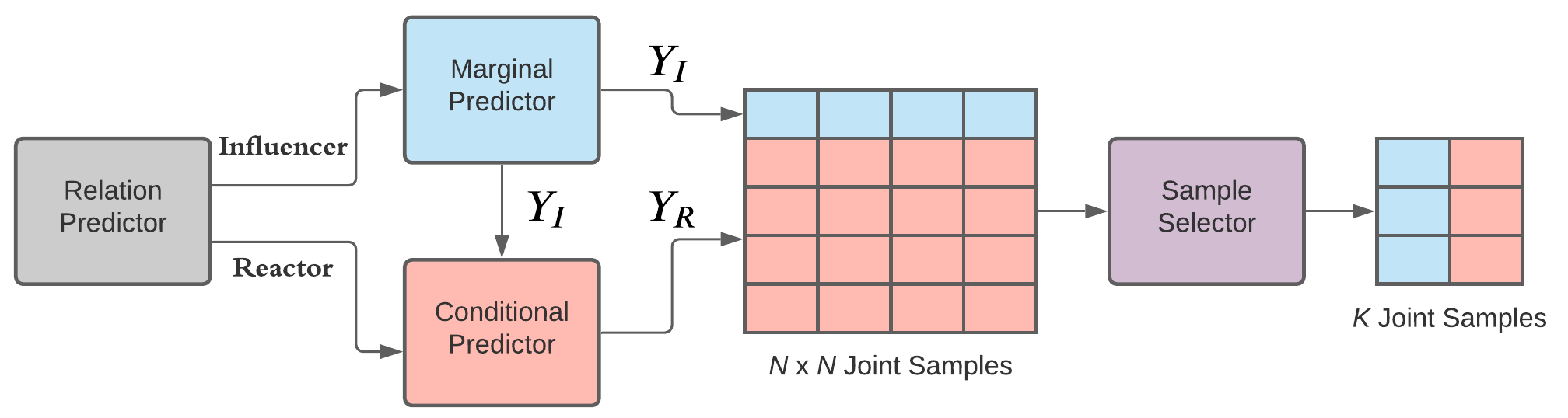}
    \caption{Overview of \modelname. The \emph{relation predictor} predicts influencer-reactor relations for interacting agents. The \emph{marginal predictor} generates marginal predictions for the influencer. The \emph{conditional predictor} generates predictions for the reactor, conditioned on each influencer trajectory. The \emph{sample selector} chooses a subset of representative joint samples as output. }
    \label{fig:model}
\vspace{-5mm}
\end{figure*}

\vspace{-1mm}
\subsection{Interactive Trajectory Prediction}
\vspace{-1mm}
Predicting scene compliant trajectories for multiple agents remains an open question due to its complexity. Early work leverages hand-crafted interaction models, such as social forces \cite{helbing1995social} and energy functions \cite{yamaguchi2011you}. These hand-crafted functions require manual tuning and have difficulties modeling highly complicated and nonlinear interactions. 
In contrast, learning-based methods achieve better accuracy by learning interactions from realistic driving data: \cite{alahi2016social,gupta2018social} utilize social pooling mechanisms to capture social influences from neighbor agents to predict interactive pedestrian trajectories in crowded scenes; \cite{mohamed2020social,casas2020spagnn,casas2020implicit,salzmann2020trajectron++} build a graph neural network (GNN) to learn the agent-to-agent interactions; \cite{kosaraju2019social,tang2019multiple,li2020end,kamra2020mapred2,ngiam2021scene} leverage attention and transformer mechanisms to learn multi-agent interaction behaviors. 
\rev{
In this work, we build a sparse graph with directed edges representing dependencies between agent nodes, but our approach differs from existing graph-based models in a few ways. First, it adopts explicit influencer-reactor relations and offers better interpretability in agent interactions. 
Second, M2I predicts scene compliant trajectories through marginal and conditional predictors to afford better computational efficiency.
Third, it utilizes the future trajectory of influencer agents to predict conditional behaviors for the reactors for better accuracy. This also allows M2I to be used for counterfactual reasoning in simulation applications by varying influencer trajectories.
}

Existing marginal prediction work produces scene compliant trajectories by leveraging an auxiliary collision loss~\cite{li2020end} or a critic based on an inverse reinforcement learning framework~\cite{van2019safecritic} that discourages colliding trajectories. In this work, we focus on identifying agent relations explicitly as influencers and reactors to generate scene compliant predictions. Our work is relevant to~\cite{lee2019joint,kumar2020interaction} that predicts interacting types before predicting scene compliant trajectories, but we further exploit the structure of the decoupled relations and the influence of low-level influencer trajectories, as opposed to only providing the high-level interaction labels as the input to the trajectory predictor. 

\subsection{Conditional Trajectory Prediction}
Conditional prediction approaches study the correlations between future agent trajectories, by predicting trajectories conditioned on the future trajectory of another agent~\cite{khandelwal2020if,salzmann2020trajectron++,tolstaya2021identifying}. These approaches often rely on the future trajectory of the autonomous vehicle or a robot whose future plan is known to the predictor. Our work goes beyond by conditioning on the future trajectory of another agent to be predicted. Despite the prediction errors of the conditioned agent, we show that our model outperforms marginal predictors that do not account for the interactive correlations.

\vspace{-2mm}
\section{Approach}
\vspace{-1mm}
In this section, we introduce a formal problem formulation and an overview of \modelname, followed by detailed explanations of each model used in the approach.

\subsection{Problem Formulation}
Given observed states $X = (M, S)$, including the map states $M$ and the observed states $S$ of all agents in a scene, the goal is to predict the future states of the interacting agents $Y$ up to a finite horizon $T$. We assume the interacting agents are pre-labeled in a given scene, which is available in common interactive prediction datasets such as~\cite{zhan2019interaction,ettinger2021large}. As the distribution over $Y$ is a joint distribution over multiple agents, we approximate it as the factorization over a marginal distribution and a conditional distribution:
\begin{equation}
    P(Y|X) = P(Y_I, Y_R|X) \approx P(Y_I|X) P(Y_R|X,Y_I).
\label{eq:fac}
\end{equation}

\begin{figure*}[t!]
    \centering
    \includegraphics[width=0.75\textwidth]{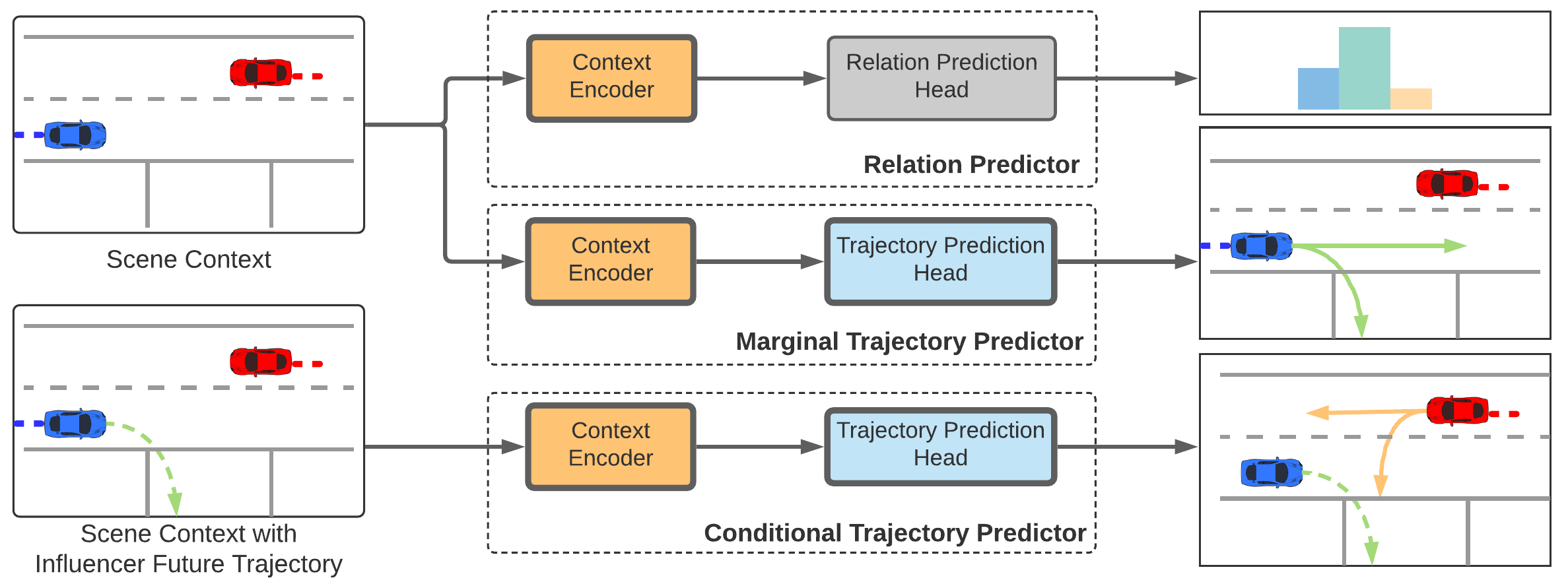}
    \caption{\modelname~includes three models that share the same context encoder. The relation predictor includes a relation prediction head to predict distribution over relation types. The marginal predictor adopts a trajectory prediction head to produce multi-modal prediction samples. The conditional trajectory predictor takes an augmented scene context input as the influencer future trajectory.}
    \label{fig:model_impl}
\vspace{-6mm}
\end{figure*}

The factorization in \cref{eq:fac} first assigns the interacting agents as the influencer $Y_I$ and the reactor $Y_R$, and decouples the joint distribution as the marginal distribution over the influencer and the conditional distribution over the reactor. This factorization allows us to reduce the complexity of learning a joint distribution to learning more tractable distributions. In the case where two agents are not interacting, the factorization can be simplified as two marginal distributions:
\begin{equation}
    P(Y|X) \approx P(Y_I|X) P(Y_R|X),
\label{eq:marg}
\end{equation}
where there is no conditional dependence between the agents. Such ind\textbf{}ependence is presumed by many marginal prediction models that predict the marginal distribution without considering other agents in the scene.

We focus on two interactive agents in this paper and aim to tackle the pairwise interactive trajectory prediction problem proposed by~\cite{ettinger2021large}. For scenarios involving more than two interactive agents, our approach can be modified by predicting the relations over all the agents and chaining multiple marginal and conditional distributions together\rev{, assuming no loopy influence:}
\rev{
\vspace{-2mm}
\begin{equation}
    P_{N>2}(Y|X) \approx \prod_{i=1}^N P(Y_i|X, \mathbf{Y}_i^{\text{inf}}),
\label{eq:marg_N}
\vspace{-2mm}
\end{equation}
where $N$ is the number of total interactive agents, and $\mathbf{Y}_i^{\text{inf}}$ is the set of influencer agents for agent $i$ predicted by the relation predictor. We refer to examples of multi-agent relation predictions in \cref{app:multi}.
}

\subsection{Model Overview}
Our proposed approach \modelname~is summarized in \cref{fig:model}. It includes a \emph{relation predictor} to predict the influencer and the reactor in a scene, a \emph{marginal predictor} to predict future trajectories of the influencer, a \emph{conditional predictor} to predict future trajectories of the reactor conditioned on the future trajectory of the influencer, and a \emph{sample selector} to select a set of representative joint prediction samples. 
Although \modelname~includes three different learned models, they share the same \emph{encoder-decoder} structure and adopt the same \emph{context encoder} to learn context information, as illustrated in \cref{fig:model_impl}. The conditional predictor takes an augmented scene context input that includes the influencer future trajectory to learn reactive behaviors for the reactor. In the following, we introduce each model with more details.

\subsection{Relation Predictor}
\label{sec:relation}
We propose a relation predictor to classify whether an interacting agent is an influencer or a reactor, based on the pass yield relation between two agents. Similar to \cite{kumar2020interaction}, we assume three types of relations: \textsc{pass}, \textsc{yield}, and \textsc{none}, and determine the relation using the following heuristics.
Given two agent future trajectories $y_1$ and $y_2$ with $T$ steps, we first compute the closest spatial distance between two agents to determine whether a pass yield relation exists:
\begin{equation}
    d_I = {\min}_{\tau_1=1}^T {\min}_{\tau_2=1}^T ||y_1^{\tau_1} - y_2^{\tau_2}||_2.
\end{equation}
If $d_I > \epsilon_d$, which is a dynamic threshold depending on the agent size, the agents never get too close to each other and thus we label the relation type as none. Otherwise, we obtain the time step from each agent at which they reach the closest spatial distance, such that:
\begin{eqnarray}
    t_1 &=& {\arg \min}_{\tau_1=1}^T {\min}_{\tau_2=1}^T ||y_1^{\tau_1} - y_2^{\tau_2}||_2, \\
    t_2 &=& {\arg \min}_{\tau_2=1}^T {\min}_{\tau_1=1}^T ||y_1^{\tau_1} - y_2^{\tau_2}||_2.
\end{eqnarray}
When $t_1 > t_2$, we define that agent 1 yields to agent 2, as it takes longer for agent 1 to reach the interaction point. Otherwise, we define that agent 1 passes agent 2. 

After labeling the training data with three interaction types, we propose an encoder-decoder-based model to classify an input scenario into a distribution over these types. 
As shown in \cref{fig:model_impl}, the relation predictor model consists of a context encoder that extracts the context information, including the observed states of the interacting agents and nearby agents and map coordinates, into a hidden vector, as well as a relation prediction head that outputs the probability over each relation type. There is a rich set of literature on learning context information from a traffic scene, such as~\cite{cui2019multimodal,gao2020vectornet,liang2020learning,gilles2021home}. Our model could utilize any existing context encoder thanks to its modular design, and we defer a detailed explanation of our choice in~\cref{sec:exp}. The relation prediction head consists of one layer of multi-layer perceptron (MLP) to output the probability logits over each relation.

The loss to train the relation predictor is defined as:
\begin{equation}
    \mathcal{L}_{\text{relation}} = \mathcal{L}_{ce}(R, \hat{R}),
\end{equation}
where $\mathcal{L}_{ce}$ is the cross entropy loss, $R$ is the predicted relation distribution, and $\hat{R}$ is the ground truth relation.

Given the predicted relation, we can assign each agent as an influencer or a reactor. If the relation is none, both agents are influencer, such that their future behaviors are independent of each other, as in \cref{eq:marg}. If the relation is agent 1 yielding to agent 2, we assign agent 1 as the reactor and agent 2 as the influencer. If the relation is agent 1 passing agent 2, we flip the influencer and reactor labels.

\subsection{Marginal Trajectory Predictor}
We propose a marginal trajectory predictor for the influencer based on an encoder-decoder structure, as shown in~\cref{fig:model_impl}, which is widely adopted in the trajectory prediction literature~\cite{zhao2020tnt,gilles2021home,ettinger2021large}. The predictor utilizes the same context encoder as in~\cref{sec:relation}, and generates a set of prediction samples associated with confidence scores using a trajectory prediction head. Although our approach can take an arbitrary prediction head, we focus on an anchor-free goal-based prediction head because of its outstanding performance in trajectory prediction benchmarks, and defer a detailed explanation in \cref{sec:exp}.

\subsection{Conditional Trajectory Predictor}
The conditional trajectory predictor is similar to the marginal predictor, except that it takes an augmented scene context that includes the future trajectory of the influencer, as shown in~\cref{fig:model_impl}. 
This allows the features of the influencer future trajectory to be extracted and learned in the same way as other context features. The encoded scene feature is used by the trajectory prediction head, which shares the same model as in the marginal predictor, to produce multi-modal prediction samples.

\subsection{Sample Selector}
Given the predicted relations of the influencer and the reactor, we predict $N$ samples with confidence scores (or probabilities) for the influencer using the marginal predictor, and for each influencer sample, we predict $N$ samples for the reactor using the conditional predictor. The number of joint samples is thus $N^2$, and the probability of each joint sample is a product of the marginal probability and the conditional probability. We further reduce the size of the joint samples to $K$ as evaluating each prediction sample for downstream tasks such as risk assessment can be expensive~\cite{wang2020fast}. In \modelname, we select the $K$ samples from $N^2$ candidates with the highest joint likelihoods.

\subsection{Inference}
At inference time, we generate the joint predictions following the procedure illustrated in \cref{fig:model}. First, we call the relation predictor and choose the interaction relation with the highest probability. Second, for the predicted influencer, we generate $N$ trajectory samples using the marginal predictor. Third, for each influencer sample, we generate $N$ samples for the predicted reactor using the conditional predictor. Fourth, we use the sample selector to select $K$ representative samples from $N^2$ candidates. In the case where the predicted relation is none, we use the marginal predictor for both agents to obtain $N^2$ trajectory pairs, and follow the same sample selection step.  

\section{Experiments}
\label{sec:exp}
In this section, we introduce the dataset benchmark and details of the model, followed by a series of experiments to demonstrate the effectiveness of \modelname.

\subsection{Dataset}
We train and validate \modelnamespace in the Waymo Open Motion Dataset (WOMD), a large-scale driving dataset collected from realistic traffic scenarios. We focus on the interactive prediction task to predict the joint future trajectories of two interacting agents for the next 8 seconds \rev{with 80 time steps}, given the observations, including 1.1 seconds of agent states \rev{with 11 time steps that may include missing observations} and the map state. The dataset includes 204,166 scenarios in the training set and 43,479 examples in the validation set. \rev{The dataset provides labels on which agents are likely to interact, yet it does not specify how they interact.} During training, we pre-label the interaction type \rev{(yield, pass, or none)} of the interacting agents according to \cref{sec:relation}.

\subsection{Metrics}
We follow the WOMD benchmark by using the following metrics:
\textbf{minADE} measures the average displacement error between the ground truth future joint trajectory and the closest predicted sample out of $K=6$ joint samples. This metric is widely adopted since~\cite{gupta2018social} to measure the prediction error against a multi-modal distribution.
\textbf{minFDE} measures the final displacement error between the ground truth end positions in the joint trajectory and the closest predicted end positions from $K$ joint samples.
\textbf{Miss rate (MR)} measures the percentage of none of the $K$ joint prediction samples are within a given lateral and longitudinal threshold of the ground truth trajectory. The threshold depends on the initial velocity of the predicted agents. More details are described in~\cite{ettinger2021large}.
\textbf{Overlap rate (OR)} \rev{measures the level of scene compliance as} the percentage of the predicted trajectory of any agent overlapping with the predicted trajectories of other agents. This metric only considers the most likely joint prediction sample. A lower overlap rate indicates the predictions are more scene compliant. In this paper, we slightly modify the metric definition compared to the original version of WOMD, which considers the overlapping among all objects including the ones not predicted, so that we can measure directly the overlapping between predicted agents.
\textbf{Mean average precision (mAP)} measures the area under the precision-recall curve of the prediction samples given their confidence scores. Compared to minADE/minFDE metrics that are only measured against the best sample regardless of its score, mAP measures the quality of confidence score and penalizes false positive predictions~\cite{ettinger2021large}. It is the \emph{official} ranking metric used by WOMD benchmark and we refer to~\cite{ettinger2021large} for the implementation.

\subsection{Model Details}
We present the detailed implementation of our model and training procedure in the following sections.

\vspace{-2mm}
\subsubsection{Context Encoder}
\vspace{-2mm}
The context encoder leverages both vectorized and rasterized representations to encode traffic context. Vectorized representation takes the traffic context, including observed agent states and map states, as vectors. It is efficient at covering a large spatial space. Rasterized representation draws traffic context on a single image with multiple channels and excels at capturing geometrical information. Both representations have achieved top performance in trajectory prediction benchmarks such as Argoverse and WOMD~\cite{chang2019argoverse,ettinger2021large,gao2020vectornet,gilles2021home,gu2021densetnt}. 

In \modelname, we use the best of both worlds. First, we leverage a vector encoder based on VectorNet~\cite{gao2020vectornet} that takes observed agent trajectories and lane segments as a set of polylines. Each polyline is a set of vectors that connect neighboring points together. For each polyline, the vector encoder runs an MLP to encode the feature of vectors within the polyline and a graph neural network to encode their dependencies followed by a max-pooling layer to summarize the feature of all the vectors. The polyline features, including agent polyline features and map polyline features, are processed by cross attention to obtain the final agent feature that includes information on the map and nearby agents. We refer to~\cite{gao2020vectornet} for detailed implementations.

In addition to encoding the vectorized feature, we utilize a second encoder to learn features from a rasterized representation. Following~\cite{gilles2021home}, we first rasterize the input states into an image with 60 channels, including the position of the agents at each past time frame with the map information.
The size of the image is $224 \times 224$ and each pixel represents an area of $1\mathrm{m} \times 1\mathrm{m}$. We run a pre-trained VGG16~\cite{simonyan2014very} model as the encoder to obtain the rasterized feature. The output of the context encoder is a concatenation of the vectorized feature and the rasterized feature.

\textbf{Conditional Context Encoder} The context encoder in the conditional trajectory predictor processes the additional influencer future trajectory in the following ways. First, the future trajectory is added to the vectorized representation as an extra vector when running VectorNet. In parallel, we create extra 80 channels on the rasterized representation and draw the $(x, y)$ positions over 80 time steps in the next 8 seconds. We run the pre-trained VGG16 model to encode the augmented image, and combine the output feature with the vectorized feature as the final output.

\vspace{-2mm}
\subsubsection{Relation Prediction Head}
\vspace{-1mm}
The relation prediction head has one layer of MLP with one fully connected layer for classification. The MLP has a hidden size of 128, followed by a layer normalization layer and a ReLU activation layer. The output is the logits over three types of relations, as described in \cref{sec:relation}. 

\vspace{-2mm}
\subsubsection{Trajectory Prediction Head}
\vspace{-1mm}
The trajectory prediction head adopts DenseTNT~\cite{gu2021densetnt} to generate multi-modal future predictions for its outstanding performance in the marginal prediction benchmarks. It first predicts the distribution of the agent goals as a heatmap, through a lane scoring module that identifies likely lanes to follow, a feature encoding module that uses the attention mechanism to extract features between goals and lanes, and a probability estimation module that predicts the likelihood of goals. Next, the prediction head regresses the full trajectory over the prediction horizon conditioned on the goal. The prediction head can be combined with the context encoder and trained end-to-end.

\vspace{-2mm}
\subsubsection{Training Details}
\vspace{-1mm}
\rev{At training time, we train each model, including the relation predictor, marginal predictor, and conditional predictor, separately. Each model is trained on the training set from WOMD with a batch size of 64 for 30 epochs on 8 Nvidia RTX 3080 GPUs. The data is batched randomly. We use an Adam optimizer and a learning rate scheduler that decays the learning rate by 30\% every 5 epochs, with an initial value of 1e-3. The hidden size in the model is 128, if not specified.
We observe consistent performance over different learning rates and batch sizes.}
When training the conditional predictor, we use the teacher forcing technique by providing the ground truth future trajectory of the influencer agent.

\vspace{-1mm}
\subsection{Quantitative Results}
\vspace{-1mm}
\label{sec:quant}
\begin{table*}[t!]
    \centering
    \footnotesize
    \bgroup
    \def\arraystretch{1.05}%
    \begin{tabular}{clccc|ccc|ccc|c}
    \toprule
    & & \multicolumn{3}{c|}{Vehicle (8s)} & \multicolumn{3}{c|}{Pedestrian (8s)} & \multicolumn{3}{c|}{Cyclist (8s)} & All (8s)\\
    Set & Model & mFDE $\downarrow$ & MR $\downarrow$ & \cellcolor{gray!50}mAP $\uparrow$ & mFDE $\downarrow$ & MR $\downarrow$ & \cellcolor{gray!50}mAP $\uparrow$ & mFDE $\downarrow$ & MR $\downarrow$ & \cellcolor{gray!50}mAP $\uparrow$ & \cellcolor{gray!50}mAP $\uparrow$ \\
    \midrule
    \multirow{2}{*}{\makecell{Val.}} 
    & Waymo LSTM Baseline~\cite{ettinger2021large} & - & 0.88 & \cellcolor{gray!50}0.01 & - & 0.93 & \cellcolor{gray!50}0.02 & - & 0.98 & \cellcolor{gray!50}0.00 & \cellcolor{gray!50}0.01 \\
    & Waymo Full Baseline~\cite{ettinger2021large} & 6.07 & 0.66 & \cellcolor{gray!50}0.08 & 4.20 & 1.00 & \cellcolor{gray!50}0.00 & 6.46 & 0.83 & \cellcolor{gray!50}0.01 & \cellcolor{gray!50}0.03 \\
    & SceneTransformer~\cite{ngiam2021scene} & \textbf{3.99} & \textbf{0.49} & \cellcolor{gray!50}0.11 & \textbf{3.15} & 0.62 & \cellcolor{gray!50}\textbf{0.06} & \textbf{4.69} & \textbf{0.71} & \cellcolor{gray!50}\textbf{0.04} & \cellcolor{gray!50}0.07 \\
    & Baseline Marginal &  6.26 & 0.60 & \cellcolor{gray!50}0.16 & 3.59 & 0.63 & \cellcolor{gray!50}0.04 & 6.47 & 0.76 & \cellcolor{gray!50}0.03 & \cellcolor{gray!50}0.07 \\
    & Baseline Joint &  11.31 & 0.64 & \cellcolor{gray!50}0.14 & 3.44 & 0.93 & \cellcolor{gray!50}0.01 & 7.16 & 0.82 & \cellcolor{gray!50}0.01 & \cellcolor{gray!50}0.05 \\
    & \modelname &  5.49 & 0.55 & \cellcolor{gray!50}\textbf{0.18} & 3.61 & \textbf{0.60} & \cellcolor{gray!50}\textbf{0.06} & 6.26 & 0.73 & \cellcolor{gray!50}\textbf{0.04} & \cellcolor{gray!50}\textbf{0.09} \\
    \midrule
    \midrule
    \multirow{2}{*}{\makecell{Test}} 
    & Waymo LSTM Baseline~\cite{ettinger2021large}  & 12.40 & 0.87 & \cellcolor{gray!50}0.01 & 6.85 & 0.92 & \cellcolor{gray!50}0.00 & 10.84 & 0.97 & \cellcolor{gray!50}0.00 & \cellcolor{gray!50}0.00 \\
    & HeatIRm4~\cite{mo2021multi}  & 7.20 & 0.80 & \cellcolor{gray!50}0.07 & 4.06 & 0.80 & \cellcolor{gray!50}0.05 & 6.69 & 0.85 & \cellcolor{gray!50}0.01 & \cellcolor{gray!50}0.04 \\
    & AIR$^2$~\cite{wu2021air}  & 5.00 & 0.64 & \cellcolor{gray!50}0.10 & 3.68 & 0.71 & \cellcolor{gray!50}0.04 & 5.47 & 0.81 & \cellcolor{gray!50}\textbf{0.04} & \cellcolor{gray!50}0.05 \\
    & SceneTransformer~\cite{ngiam2021scene}  & \textbf{4.08} & \textbf{0.50} & \cellcolor{gray!50}0.10 & \textbf{3.19} & 0.62 & \cellcolor{gray!50}0.05 & \textbf{4.65} & \textbf{0.70} & \cellcolor{gray!50}\textbf{0.04} & \cellcolor{gray!50}0.06 \\
    & \modelname  &  5.65 & 0.57 & \cellcolor{gray!50}\textbf{0.16} & 3.73 & \textbf{0.60} & \cellcolor{gray!50}\textbf{0.06} & 6.16 & 0.74 & \cellcolor{gray!50}0.03 & \cellcolor{gray!50}\textbf{0.08} \\
    \bottomrule
    \end{tabular}%
    \egroup
    \caption{\underline{Joint} metrics on the interactive validation and test set. The best performed metrics are bolded and the grey cells indicate the ranking metric used by the WOMD benchmark. \modelname~ outperforms both Waymo baselines and challenge winners. Compared to the current state-of-the art model SceneTransformer, it improves the mAP metric by a large margin over vehicles and all agents, demonstrating its advantage in learning a more accurate probability distribution and producing fewer false positive predictions.}
    \label{tab:baselines}
\vspace{-4mm}
\end{table*}

In \cref{tab:baselines}, we compare our model with the following baselines, including the top ranked published models on the WOMD interaction prediction challenge leaderboard~\cite{waymo}:
\textbf{Waymo LSTM Baseline}~\cite{ettinger2021large} is the official baseline provided by the benchmark. It leverages an LSTM encoder to encode observed agent trajectories, and an MLP-based prediction head to generate multiple samples.
\textbf{Waymo Full Baseline}~\cite{ettinger2021large} is an extended version of the Waymo LSTM Baseline, by leveraging a set of auxiliary encoders to encode context information.
\textbf{SceneTransformer}~\cite{ngiam2021scene} is a transformer-based model that leverages attention to combine features across road graphs and agent interactions both spatially and temporally. The model achieves state-of-the-art performance in the WOMD benchmark in both the marginal prediction task and the interactive prediction task. 
\textbf{HeatIRm4}~\cite{mo2021multi} models the agent interaction as a directed edge feature graph and leverages an attention network to extract interaction features. It was the winner of the 2021 WOMD challenge.
\textbf{AIR$^2$}~\cite{wu2021air} adopts a marginal anchor-based model using a raster representation. The model generates joint predictions by combining marginal predictions from each agent. It achieved the top performance at the WOMD challenge.
\textbf{Baseline Marginal} is our baseline model that leverages the same marginal predictor as \modelnamespace to generate $N$ marginal prediction samples for both agents, without considering their future interactions. When combining the marginal predictions into joint predictions, we take the top $K$ marginal pairs out of $N^2$ options given their joint probabilities as the product of marginal probabilities. This is a common practice to combine marginal predictions into joint predictions, as in~\cite{casas2020implicit,ettinger2021large}.
\textbf{Baseline Joint} is our baseline model that jointly predicts the goals and trajectories for both interacting agents, using the same context encoder and the trajectory prediction head as in \modelname. As the joint goal space grows exponentially with the number of agents, we can only afford a small number of goal candidates for each agent. To ease the computational complexity, we leverage a marginal predictor to predict the top 80 goals for each agent and obtain \rev{$80\times80$} goal pairs for joint goal and trajectory prediction. As a result, this baseline trade-offs prediction accuracy with computational feasibility by using a reduced set of goals.

\vspace{-2mm}
\subsubsection{Validation Set}
\vspace{-1mm}
We present the results in the interactive validation set in the top half of~\cref{tab:baselines}, where the baseline results are reported as in~\cite{ettinger2021large,ngiam2021scene}. Our model \modelnamespace outperforms both Waymo baselines in terms of all metrics. Compared to the current state-of-the-art model SceneTransformer, \modelname~achieves a better \emph{mAP}, the official ranking metric, over vehicles, and a better miss rate over pedestrians. Although \modelname~has higher minFDE errors, it has improved the mAP over all agents (the most right column) by a large margin, meaning our model generates a more accurate distribution using its predicted confidence scores and outputs fewer false positive predictions. In addition, as our proposed approach does not assume a specific prediction model, it could leverage SceneTransformer as the context encoder to achieve better minFDE, and we defer it as future work.
When compared with our own baselines that share the same context encoder and prediction head, \modelnamespace outperforms the marginal predictor, which assumes independence between two agents, and a joint predictor, which only affords a small set of goal candidates due to computational constraints.

\vspace{-4mm}
\subsubsection{Testing Set}
\vspace{-1mm}
We show the results in the interactive test set in the bottom half of~\cref{tab:baselines}. For a fair comparison, we use the numbers reported on the official benchmark website~\cite{waymo} and only include the published models. Similar to the observations from the validation set, we observe that \modelnamespace improves mAP metrics by a large margin, compared to past WOMD interaction prediction challenge winners~\cite{mo2021multi,wu2021air} and the existing state-of-the-art model~\cite{ngiam2021scene}.

\subsection{Ablation Study}
\vspace{-1mm}
We present ablation studies on the relation predictor, conditional predictor, and generalization to other predictors.
\rev{
\vspace{-6mm}
\subsubsection{Relation Prediction}
\vspace{-1mm}
We measure the performance of our relation predictor on the validation dataset and observe an accuracy of 90.09\%.
We verify the significance of an accurate relation predictor by comparing the performance of vehicle trajectory predictions using the predicted relations and using the ground truth relations, and observe a gap of 3.05\% in terms of mAP at 8s. 
}

\begin{table}[t!]
\vspace{1mm}
    \centering
    \footnotesize
    \bgroup
    \def\arraystretch{1.05}%
    \begin{tabular}{lcccc}
    \toprule
    Model & minADE $\downarrow$ & minFDE $\downarrow$ & MR $\downarrow$ & mAP $\uparrow$ \\
    \midrule
    \modelnamespace Marginal & 1.70 & 3.45 & 0.23 & 0.30 \\
    \modelnamespace Conditional GT & \textbf{1.46} & \textbf{2.43} & \textbf{0.12} & \textbf{0.41} \\
    \modelnamespace Conditional P1 & 1.75 & 3.49 & 0.25 & 0.26 \\
    \bottomrule
    \end{tabular}%
    \egroup
    \caption{Comparison between the marginal predictor and the conditional predictor over \underline{marginal} metrics for vehicle reactors at 8s.}
    \label{tab:conditional}
\vspace{-4mm}
\end{table}

\begin{figure*}[t!]
    \centering
    \includegraphics[width=0.9\textwidth]{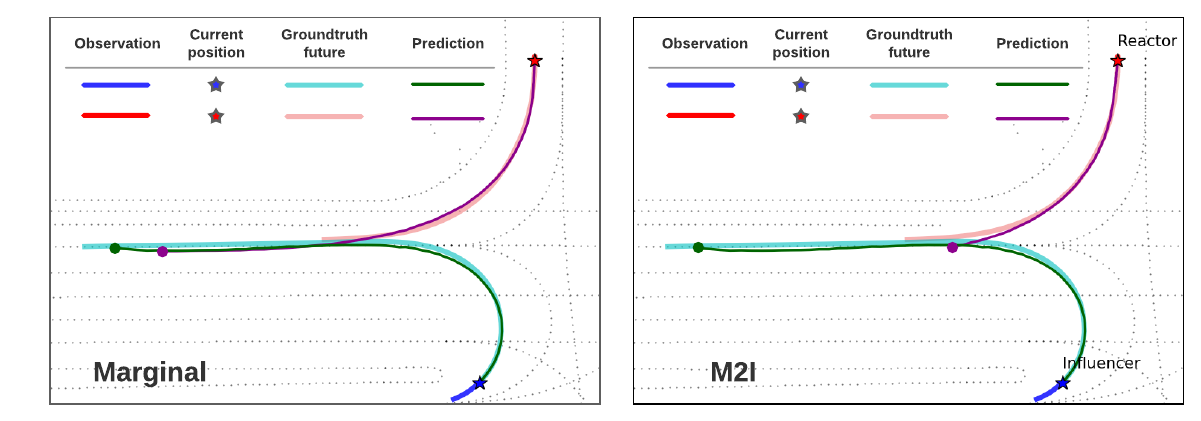}
    \caption{Example prediction using Baseline Marginal (left) and \modelnamespace(right). The marginal predictor produces overlapping and inaccurate predictions. \modelname~successfully identifies the influencer and reactor (the predicted relation type is annotated next to the current position of each agent) in a challenging interactive scene and achieves better prediction accuracy and scene compliance.}
    \label{fig:qualitative}
\vspace{-4mm}
\end{figure*}

\vspace{-2mm}
\subsubsection{Conditional Prediction}
\vspace{-1mm}
We validate the effectiveness of our conditional predictor by comparing its performance against the marginal predictor (\emph{\modelnamespace Marginal}) for vehicle reactor trajectory prediction. The results are summarized in \cref{tab:conditional}. When the conditional predictor takes the ground truth future trajectory of the influencer agent (c.f. \emph{\modelnamespace Conditional GT}), it generates predictions for the reactor agent with better performance across all metrics. This validates our hypothesis on the dependence between the influencer trajectory and the reactor trajectory.
As the ground truth trajectories are not available at inference time, we present the prediction results when the conditional predictor takes the best predicted influencer trajectory as \emph{\modelnamespace Conditional P1}. It is not surprising to see that the performance is inferior to the marginal predictor results, due to errors in influencer prediction. However, as we show in \cref{tab:baselines}, our model is able to outperform the marginal baseline model by including more than one sample from the influencer and selecting the most likely joint samples.

\begin{table}[t!]
    \centering
    \footnotesize
    \bgroup
    \def\arraystretch{1.05}%
    \begin{tabular}{lcccc}
    \toprule
    Model & minADE $\downarrow$ & minFDE $\downarrow$ & OR $\downarrow$ & mAP $\uparrow$ \\
    \midrule
    TNT Marginal & 3.43 & 8.72 & 0.42 & 0.10 \\
    TNT Joint & 5.30 & 14.07 & 0.34 & 0.13 \\
    TNT \modelname & \textbf{3.38} & \textbf{8.46} & \textbf{0.20} & \textbf{0.14} \\
    \bottomrule
    \end{tabular}%
    \egroup
    \caption{\underline{Joint} metrics on the interactive validation set for vehicles at 8s. We replace the context encoder and the prediction head in \modelnamespace and baselines with a different model. We observe a similar trend in performance improvement, especially over OR and mAP, which validates the generalizability of our proposed approach.}
    \label{tab:tnt}
\vspace{-4mm}
\end{table}

\vspace{-2mm}
\subsubsection{Generalizing to Other Predictors}
\vspace{-1mm}
We demonstrate that our proposed approach can be extended to other existing predictor models to validate its generalizability. In this experiment, we replace the context encoder with VectorNet~\cite{gao2020vectornet} and the prediction head with TNT~\cite{zhao2020tnt}, which is an anchor-based goal-conditioned prediction model, and obtain a variant of \modelnamespace named \emph{TNT \modelname}. 
We compare this variant with a marginal predictor baseline (\emph{TNT Marginal}) and a joint predictor baseline (\emph{TNT Joint}) using the same VectorNet and TNT backbones. 

The results, summarized in \cref{tab:tnt}, show that our approach consistently improves all metrics, especially OR and mAP, by a large margin when using a different predictor model. The improvements indicate that our proposed approach generalizes to other predictors and generates scene compliant and accurate future trajectories.

\subsection{Qualitative Results}
We present a challenging interactive scenario\footnote{More examples can be found in \cref{sec:additional_ex}.} in \cref{fig:qualitative}, and visualize the most likely prediction sample from a marginal baseline and \modelname. In this scenario, the red agent is yielding to the blue agent who is making a U-turn. The marginal predictor on the left fails to capture the interaction and predicts overlapping trajectories. On the other hand, \modelname~successfully identifies the underlying interaction relation, and predicts an accurate trajectory for the influencer and an accurate reactor trajectory that reacts to the predicted influencer trajectory. As a result, \modelname~achieves better prediction accuracy and scene compliance.

\vspace{2mm}
\section{Conclusion}
In conclusion, we propose a simple but effective joint prediction framework \modelname~through marginal and conditional predictors, by exploiting the factorized relations between interacting agents.
\modelname~uses a modular encoder-decoder architecture, allowing it to choose from a variety of context encoders and prediction heads. 
Experiments on the interactive Waymo Open Motion Dataset benchmark show that our framework achieves state-of-the-art performance. 
In the ablation study, we show the generalization of our framework using a different predictor model. 

\vspace{4mm}
\textbf{Limitations} We identify the following limitations.
First, there exists a gap when comparing our model to the state-of-the-art in terms of the minFDE metric, indicating that our approach still has room for improvement. Thanks to its modular design, we plan to extend \modelname~to use SceneTransformer~\cite{ngiam2021scene} as the context encoder and fill the gap. Second, the performance of \modelname~heavily depends on the size of interactive training data, especially when training the relation predictor and the conditional trajectory predictor. Looking at \cref{tab:baselines}, we see that our approach improves the mAP metrics by a large margin on vehicles because of sufficient vehicle interactions in the training data, but the improvement is more negligible over the other two types due to lack of interactive scenarios involving pedestrians and cyclists. 
\rev{
Finally, M2I assumes no mutual influence between interacting agents, allowing it to decouple joint agent distributions into marginal and conditional distributions.
While we have observed an obvious influencer according to our heuristics in almost all the interactive scenarios in the Waymo Open Motion Dataset, we defer predicting for more complicated scenarios involving mutual influence (and loopy influence for more than two agents) as future work.
}

\newpage
{\small
\bibliographystyle{ieee_fullname}
\bibliography{reference}
}

\newpage
\appendix
\section*{Appendix}
\renewcommand{\thesubsection}{\Alph{subsection}}
\subsection{Additional Experiment Details}

In this section, we introduce additional details on filtering interactive training data, training the baseline joint predictor, and training by agent types. 

\subsubsection{Filtering Interactive Training Data}
The Waymo Open Motion Dataset only provides interactive scenarios in its validation set and testing set. To filter the interactive scenario in the training set, we implement a script to identify scenarios that include 2 interacting agents based on the \emph{objects\_of\_interest} mask provided in the data. The script is provided in the source code.

\subsubsection{Baseline Joint Predictor}
We train the \emph{Baseline Joint} predictor described in \cref{sec:quant} as follows. First, we predict the distribution of goals for each interacting agent as a heatmap, according to~\cite{gu2021densetnt}. Second, we select the top 80 goals based on the predicted probability for each agent. Third, we combine the selected goals into 6400 goal pairs and run each goal pair feature, including $(x, y)$ positions for both goals, through a 2-layer MLP with a hidden size of 128 followed by a normalization layer and a ReLU activation layer. Fourth, we run a fully connected layer to predict the probability logit for each goal pair, and train the joint goal prediction model through the following loss:
\begin{equation}
    \mathcal{L}_J = \mathcal{L}_{ce}(J, \hat{J}),
\end{equation}
where $\mathcal{L}_{ce}$ is the cross entropy loss, $J$ is the predicted goal pair distribution, and $\hat{J}$ is the index of the goal pair out of all candidates that is the closest to the ground truth goal pair in terms of Euclidean distance. Given the predicted goal pairs, we train the trajectory completion model to regress the full trajectories of both interacting agents following the same procedure in~\cite{gu2021densetnt}.

\subsubsection{Training by Agent Types}
The Waymo Open Motion Dataset consists of three types of agents to predict: vehicles, pedestrians, and cyclists. As each agent type has different behavior models and the distribution is unbalanced among types (e.g. vehicle types account for 78\% of the training data), we train the marginal trajectory predictor and the conditional trajectory predictor for each agent type separately. We observe that the prediction performance over pedestrians and cyclists improves by a large margin, compared to training a single model for all agents.  

For the same reason, we train four relation predictors for vehicle-vehicle interactions, vehicle-pedestrian interactions, vehicle-cyclist interactions, and interactions that cover the remaining agent pair types, including cyclist-pedestrian, cyclist-cyclist, pedestrian-pedestrian.

\subsection{Additional Qualitative Examples}
\label{sec:additional_ex}
We present additional representative examples in a variety of interaction settings to showcase the advantage of \modelname~over the marginal baseline.

\subsubsection{Influencer Overtakes Reactor}
\begin{figure*}[h]
    \begin{minipage}{0.49\textwidth}
	    \centering
        \includegraphics[width=1.03\textwidth]{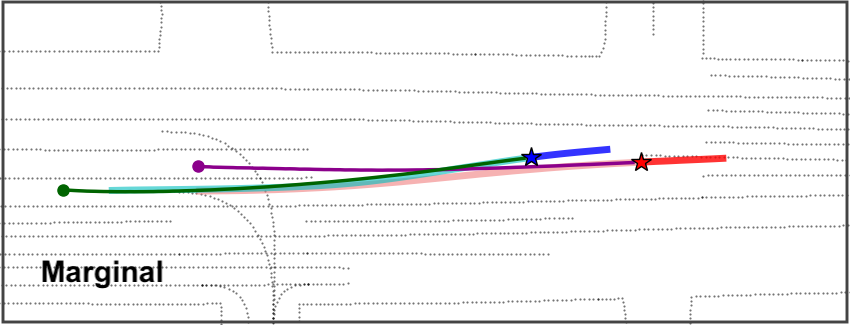}
    \end{minipage}
    \hfill
    \begin{minipage}{0.49\textwidth}
	    \centering
        \includegraphics[width=1.03\textwidth]{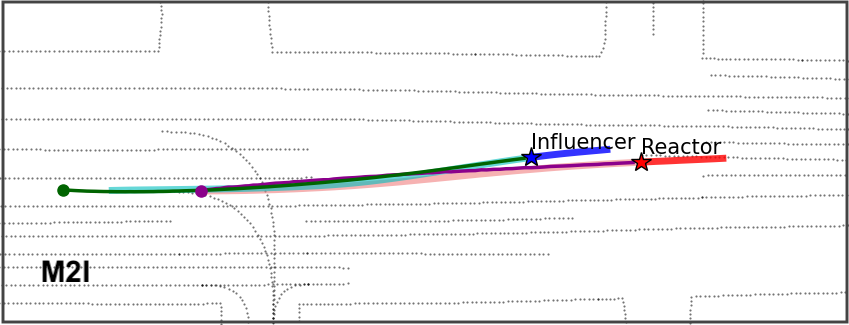}
    \end{minipage}\\
    \begin{minipage}{0.49\textwidth}
	    \centering
        \includegraphics[width=1.03\textwidth]{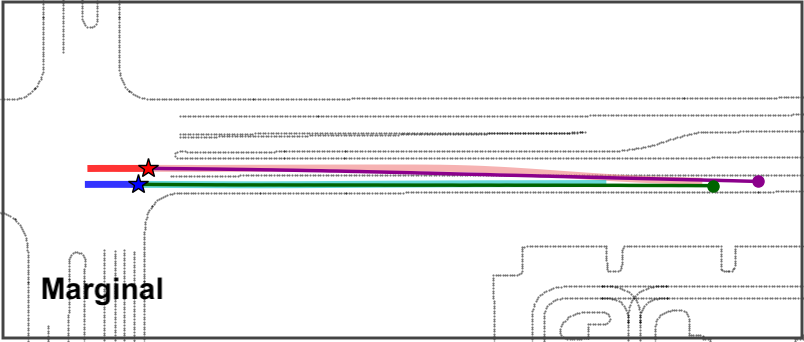}
    \end{minipage}
    \hfill
    \begin{minipage}{0.49\textwidth}
	    \centering
        \includegraphics[width=1.03\textwidth]{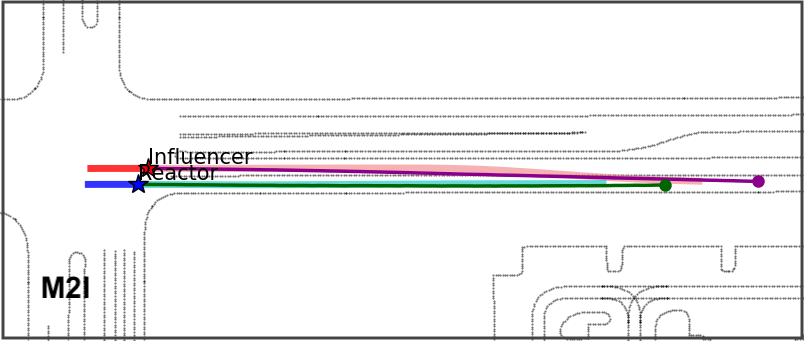}
    \end{minipage}\\
    \begin{minipage}{0.49\textwidth}
	    \centering
        \includegraphics[width=1.03\textwidth]{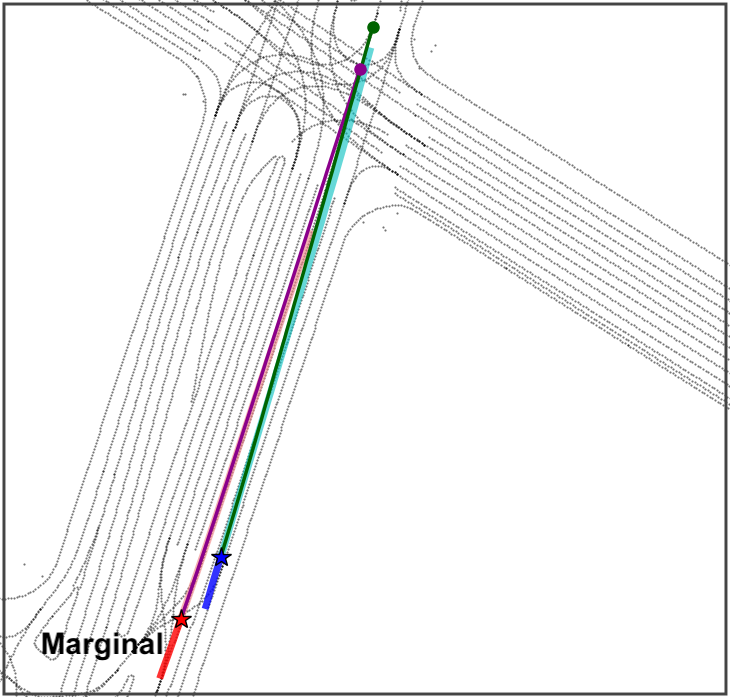}
    \end{minipage}
    \hfill
    \begin{minipage}{0.49\textwidth}
	    \centering
        \includegraphics[width=1.03\textwidth]{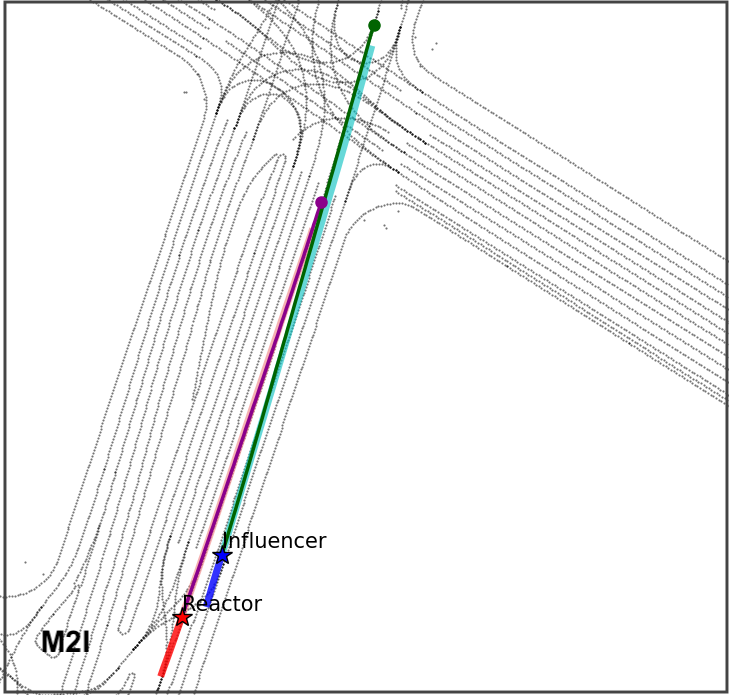}
    \end{minipage}\\
    \begin{center}
        \includegraphics[width=0.3\textwidth]{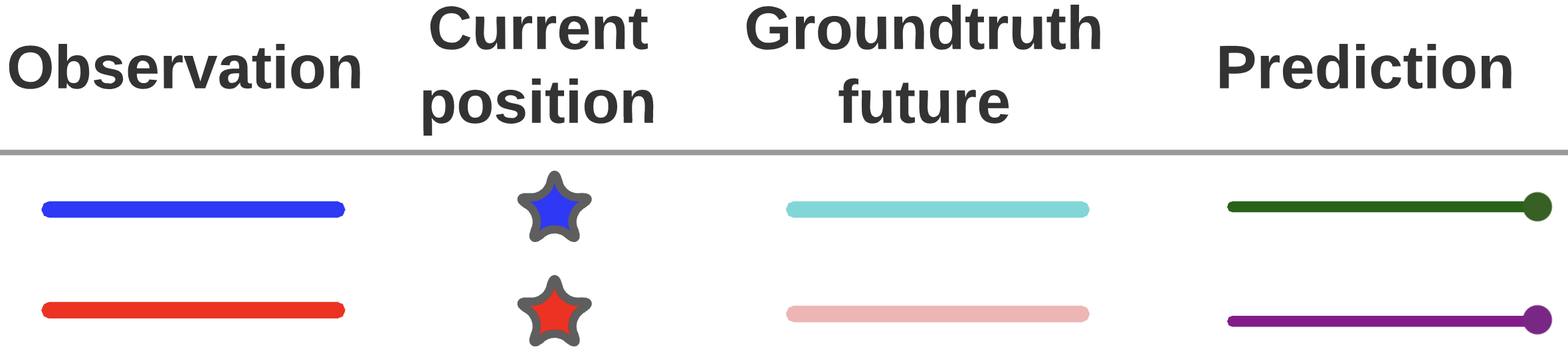}
    \end{center}
    \caption{Influencer overtakes reactor. In each example, \modelname~(right column) successfully predicts the correct relation type and improves prediction accuracy and scene compliance, while the marginal predictor (left column) predicts overlapping trajectories without considering the future interaction between agents.}
    \label{fig:qualitative_1}
\end{figure*}

In \cref{fig:qualitative_1}, we present three examples in which the influencer overtakes the reactor. In each example, \modelname~successfully predicts the correct relation type and improves prediction accuracy and scene compliance, while the marginal predictor predicts overlapping trajectories without considering the future interaction between agents.

\subsubsection{Reactor Yields to Influencer before Turning}
\begin{figure*}[h]
    \begin{minipage}{0.49\textwidth}
	    \centering
        \includegraphics[width=1.03\textwidth]{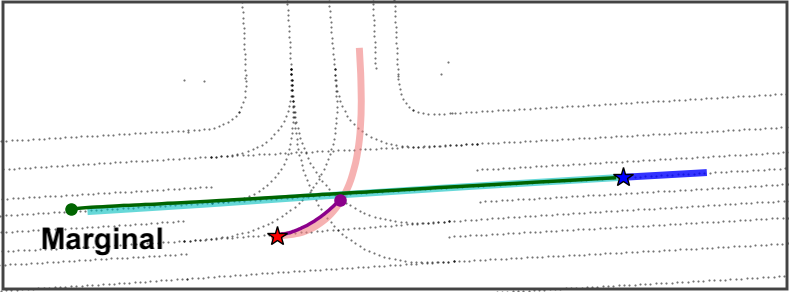}
    \end{minipage}
    \hfill
    \begin{minipage}{0.49\textwidth}
	    \centering
        \includegraphics[width=1.03\textwidth]{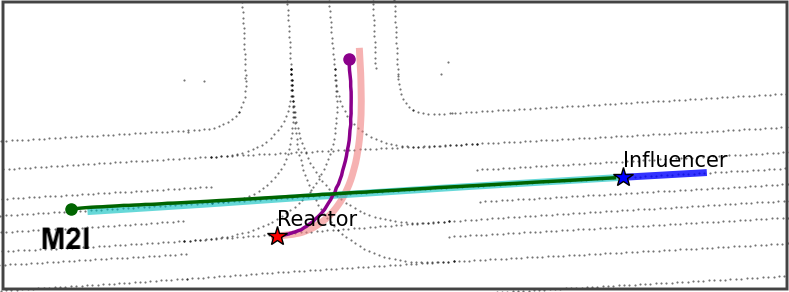}
    \end{minipage}\\
    \begin{minipage}{0.49\textwidth}
	    \centering
        \includegraphics[width=1.03\textwidth]{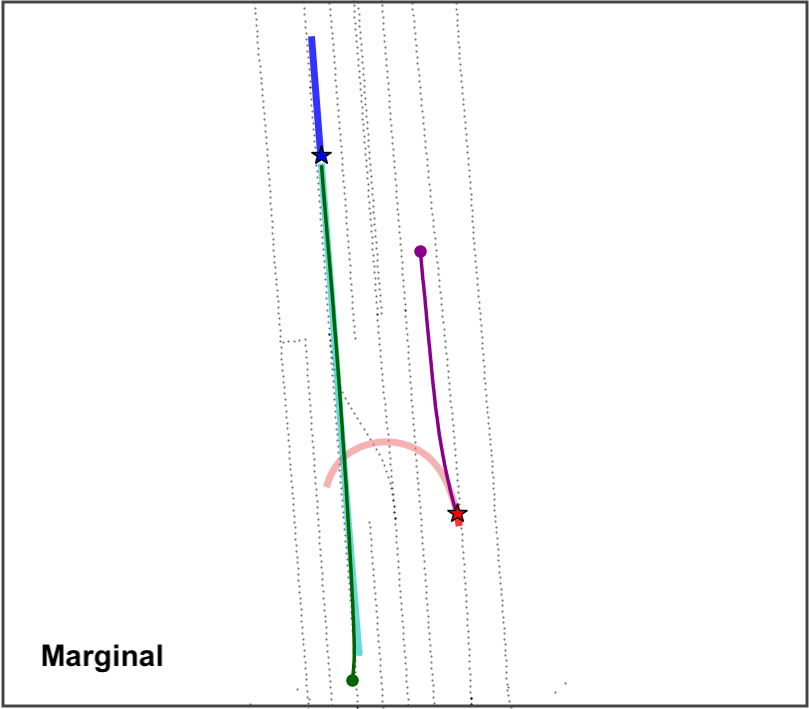}
    \end{minipage}
    \hfill
    \begin{minipage}{0.49\textwidth}
	    \centering
        \includegraphics[width=1.03\textwidth]{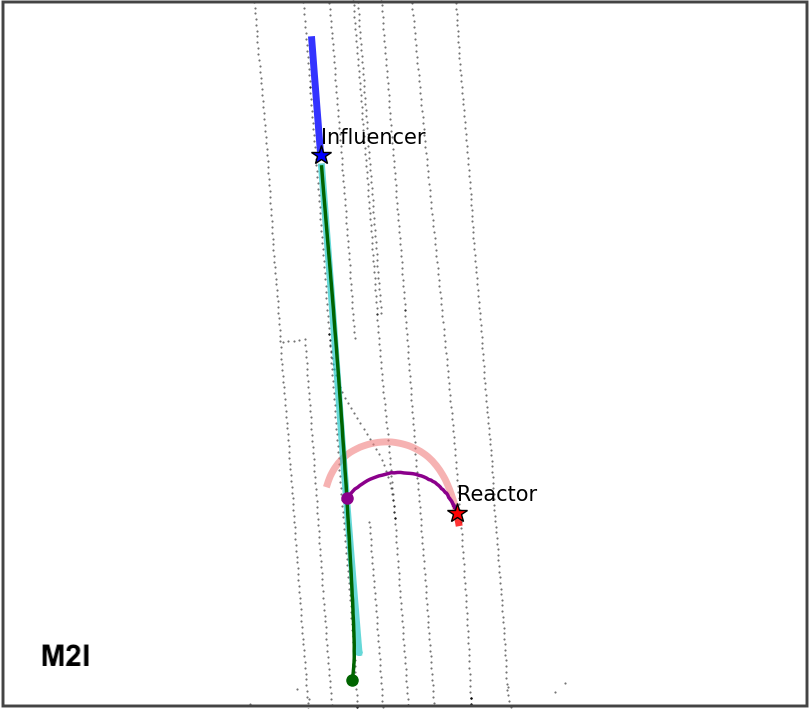}
    \end{minipage}\\
    \begin{minipage}{0.49\textwidth}
	    \centering
        \includegraphics[width=1.03\textwidth]{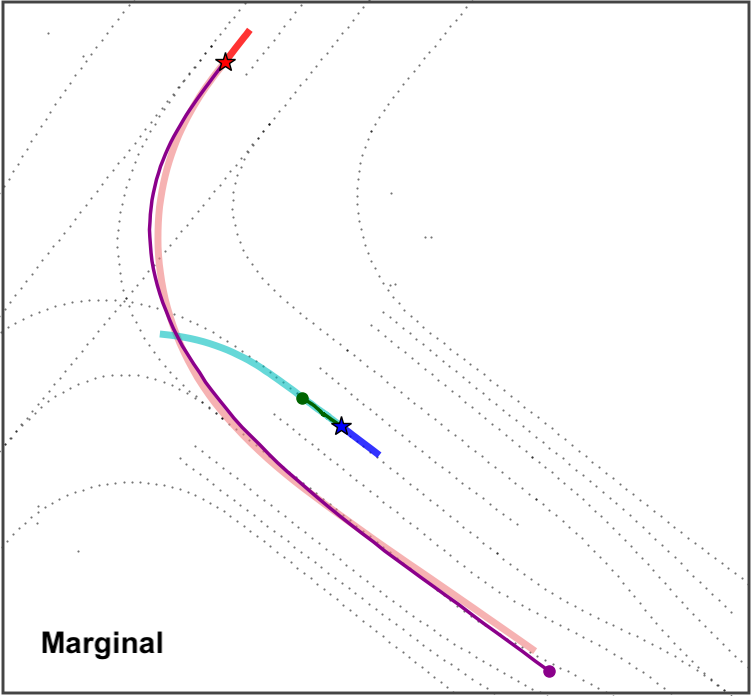}
    \end{minipage}
    \hfill
    \begin{minipage}{0.49\textwidth}
	    \centering
        \includegraphics[width=1.03\textwidth]{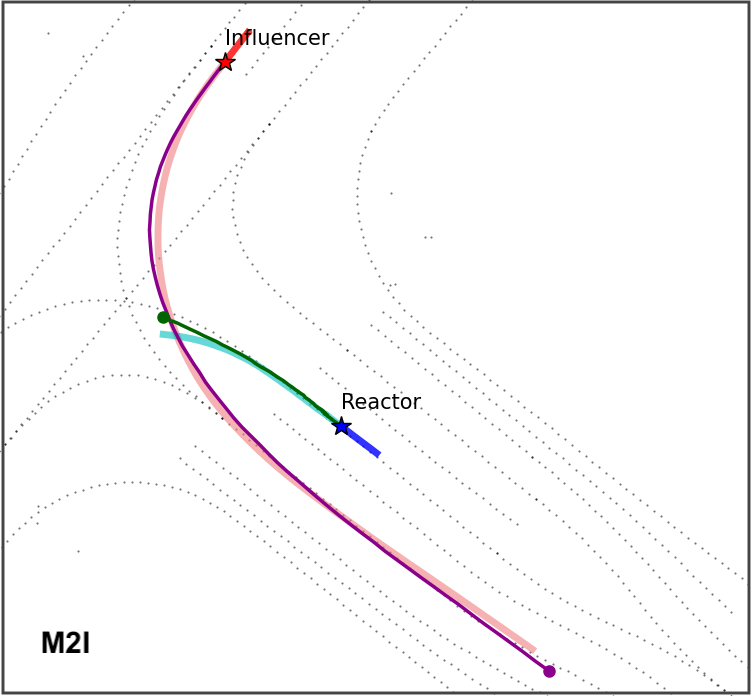}
    \end{minipage}\\
    \begin{center}
        \includegraphics[width=0.3\textwidth]{images/examples/m2i_annotation.png}
    \end{center}
    \caption{Reactor yields to influencer before turning. In each example, \modelname~(right column) successfully predicts the correct relation type and the accurate reactive trajectories for the reactor. On the other hand, the marginal predictor (left column) ignores the interaction and results in less accurate predictions.}
    \label{fig:qualitative_2}
\end{figure*}

In \cref{fig:qualitative_2}, we present three examples in which the reactor waits for the influencer to pass before turning. In each example, \modelname~successfully predicts the correct relation type and the accurate reactive trajectories for the reactor. On the other hand, the marginal predictor ignores the interaction and results in less accurate predictions.

\subsubsection{Reactor Merges behind Influencer}
\begin{figure*}[h]
\begin{minipage}{0.49\textwidth}
	    \centering
        \includegraphics[width=1.03\textwidth]{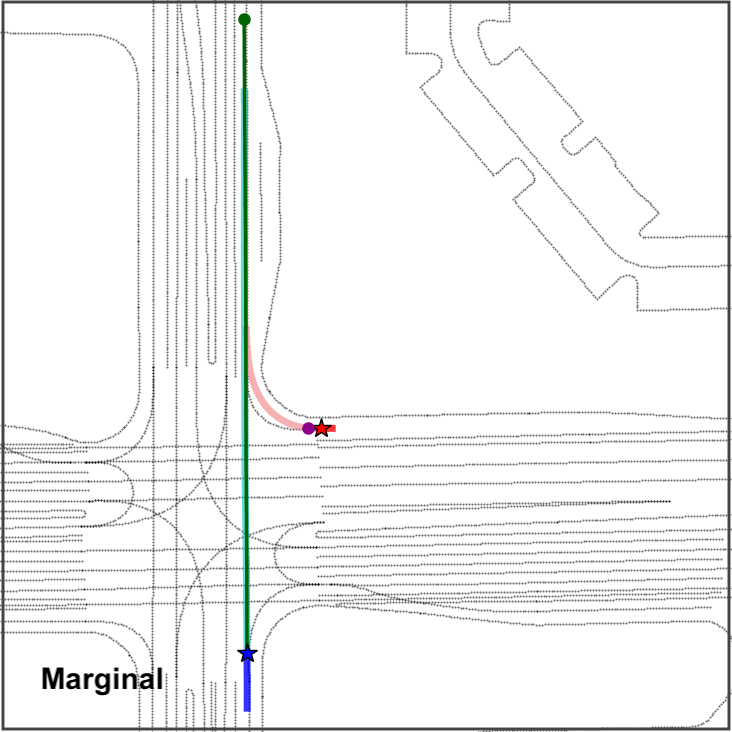}
    \end{minipage}
    \hfill
    \begin{minipage}{0.49\textwidth}
	    \centering
        \includegraphics[width=1.03\textwidth]{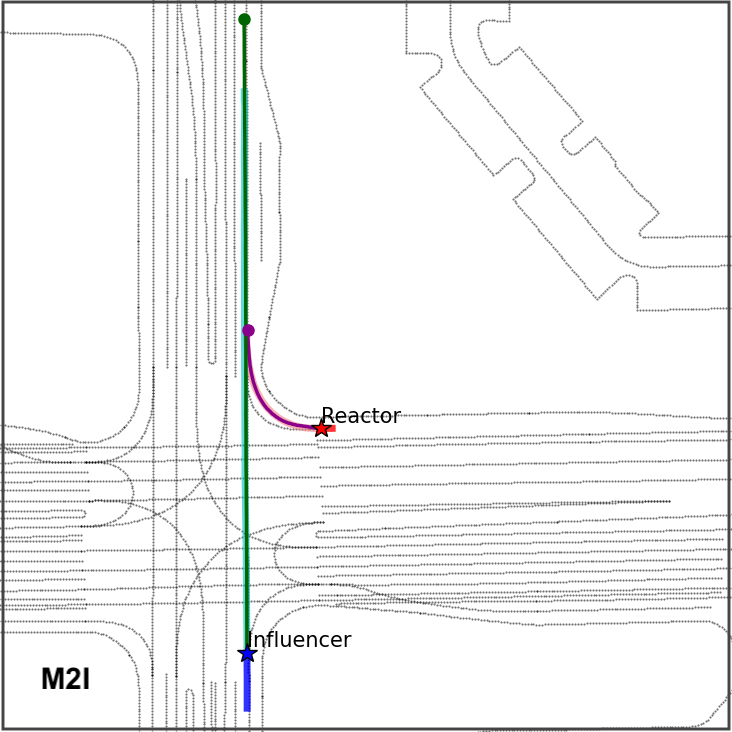}
    \end{minipage}\\
    \begin{minipage}{0.49\textwidth}
	    \centering
        \includegraphics[width=1.03\textwidth]{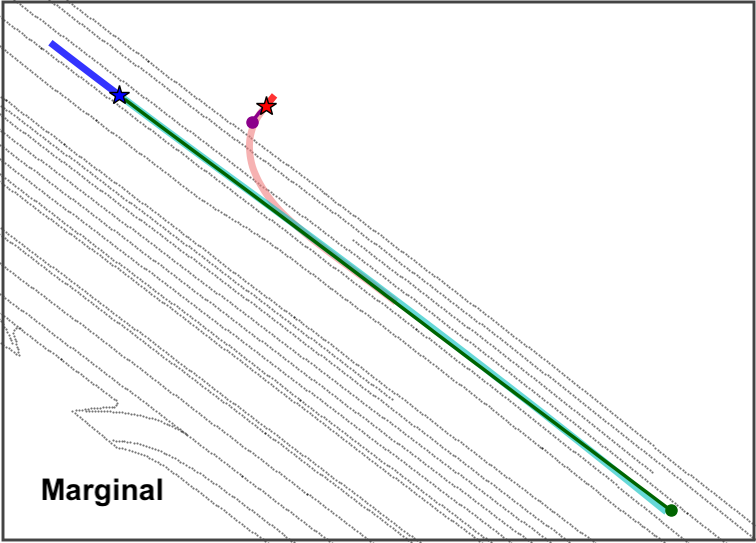}
    \end{minipage}
    \hfill
    \begin{minipage}{0.49\textwidth}
	    \centering
        \includegraphics[width=1.03\textwidth]{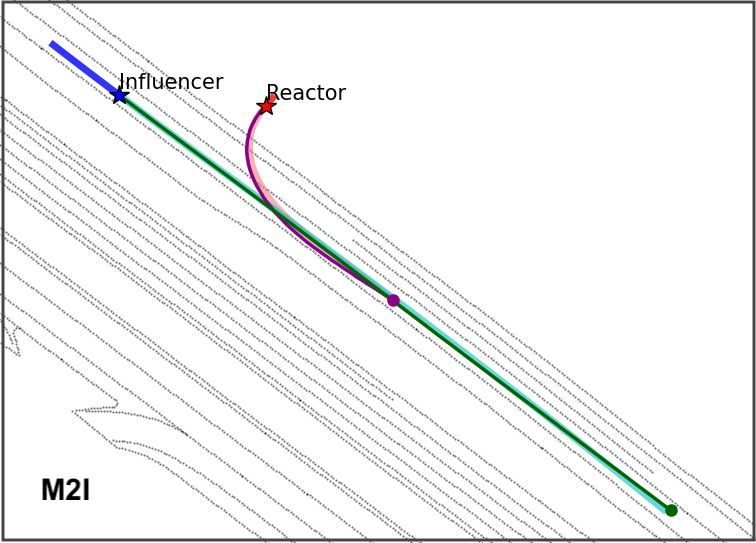}
    \end{minipage}\\
    \begin{center}
        \includegraphics[width=0.3\textwidth]{images/examples/m2i_annotation.png}
    \end{center}
    \caption{Reactor merges behind influencer. In each example, \modelname~(right column) successfully predicts the correct relation type and the accurate reactor trajectories that follow the influencer, while the marginal predictor (left column) fails to account for the interaction and predicts trajectories far away from the ground truth.}
    \label{fig:qualitative_3}
\end{figure*}

In \cref{fig:qualitative_3}, we present two examples in which the reactor merges behind the influencer after the influencer passes. In each example, \modelname~successfully predicts the correct relation type and the accurate reactor trajectories that follow the influencer, while the marginal predictor fails to account for the interaction and predicts trajectories far away from the ground truth.

\begin{figure*}[t]
  \centering
  \includegraphics[width=0.85\linewidth]{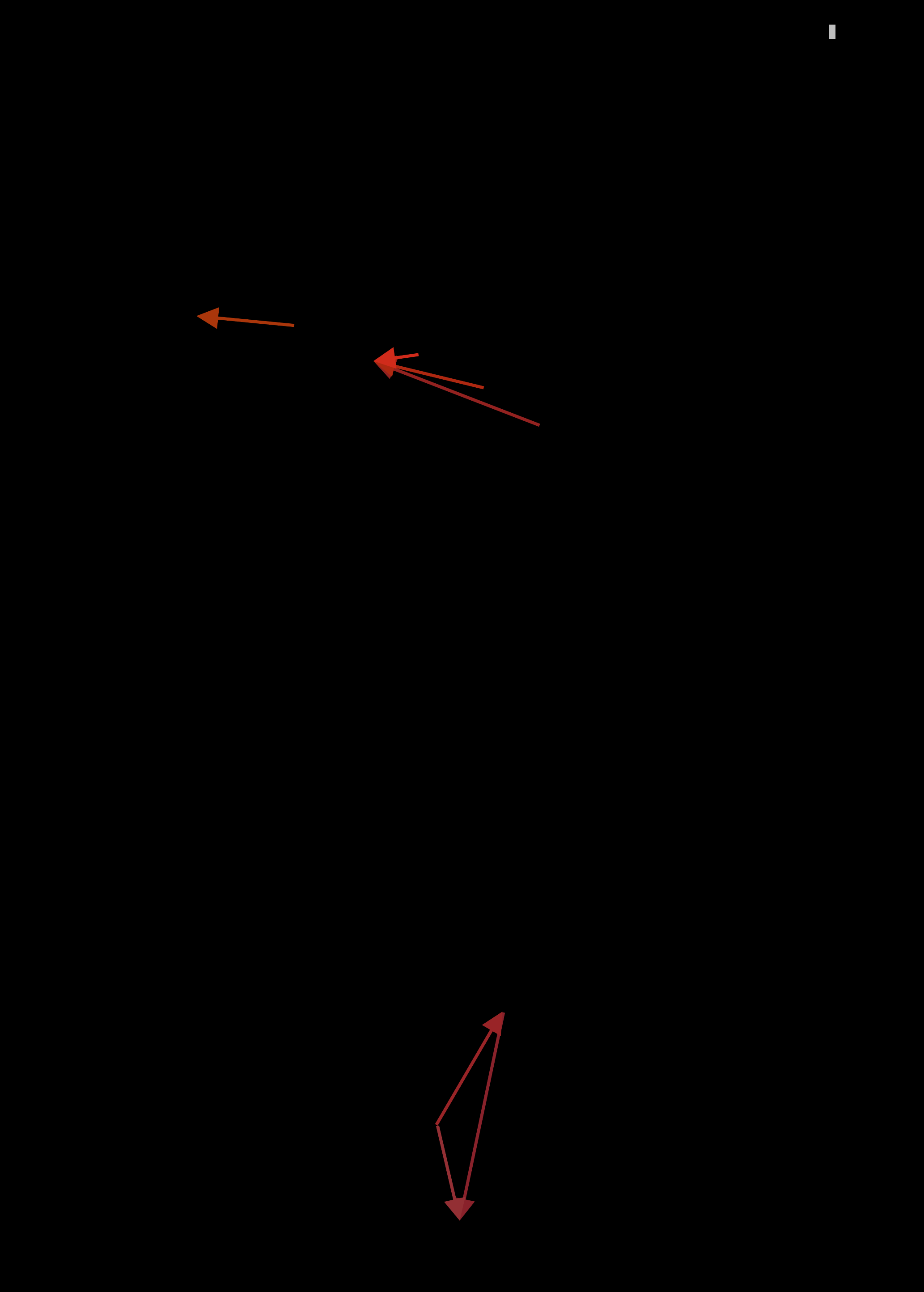}
  \caption{Examples of \modelnamespace providing scene compliant relation predictions in complex multi-agent scenarios.}
  \label{fig:multi}
\end{figure*}

\rev{
\subsection{Multi-Agent Generalization}
\label{app:multi}
We present a qualitative analysis on applying M2I to multi-agent scenarios involving more than two agents. In \cref{fig:multi}, we show two examples in which \modelnamespace provides scene compliant relation predictions in crowded traffic. Given the relation predictions, it is straightforward to predict the agent trajectories through marginal and conditional predictors, as in \cref{eq:marg_N}.
}

\end{document}